\documentclass{article}
\usepackage{ijcai20}

\usepackage{times}

\usepackage{soul}
\usepackage{url}
\usepackage[hidelinks]{hyperref}
\usepackage[utf8]{inputenc}
\usepackage[small]{caption}
\usepackage{graphicx}
\usepackage{amsmath}
\usepackage{booktabs}
\usepackage{amsmath}
\usepackage{amsthm}
\usepackage{amssymb}
\usepackage{graphicx}

\usepackage{float}
\newfloat{algorithm}{tbp}{loa}
\providecommand{\algorithmname}{Algorithm}
\floatname{algorithm}{\protect\algorithmname}

\usepackage{algorithmic}
\usepackage{algolyx}

\newtheorem{theorem}{Theorem}
\newtheorem{lem}{Lemma}

\title{Randomised Gaussian Process Upper Confidence Bound for Bayesian Optimisation}

\author{
Julian Berk\footnote{Contact Author}\and
Sunil Gupta\and
Santu Rana\And
Svetha Venkatesh\\
\affiliations
Applied Artificial Intelligence Institute\\
\emails
\{jmberk, sunil.gupta, santu.rana, svetha.venkatesh\}@deakin.edu.au
}

\begin{document}

\maketitle

\begin{abstract}
In order to improve the performance of Bayesian optimisation, we develop a modified Gaussian process upper confidence bound (GP-UCB) acquisition function. This is done by sampling the exploration-exploitation trade-off parameter from a distribution. We prove that this allows the expected trade-off parameter to be altered to better suit the problem without compromising a bound on the function's Bayesian regret. We also provide results showing that our method achieves better performance than GP-UCB in a range of real-world and synthetic problems. 
\end{abstract}

\section{Introduction}
Global optimisation is a cornerstone of modern scientific innovation. Optimisation of alloys and other materials has allowed us to create massive vehicles that are both strong and light enough to fly. Optimisation in medical science has seen us live longer and healthier lives than previously thought possible. This optimisation usually involves a trial-and-error approach of repeated experiments with different inputs to determine which input produces the most desirable output. Unfortunately, many system are expensive to sample, and the heuristic methods commonly used to select inputs are not sample-efficient. This can lead to these optimisation experiments being prohibitively costly. As such, methods that can select inputs in a sample-efficient manner can lead to faster and cheaper innovation in a wide range of fields.

\emph{Bayesian optimisation} is one of the most sample efficient methods for optimising expensive, noisy systems. It has shown excellent performance on a range of practical problems, including problems in biomedical science \cite{turgeon2016cognitive,gonzalez2015bayesian}, materials science \cite{li2017rapid,ju2017designing}, and machine learning \cite{Snoek2012,klein2017fast,xia2017boosted}.  It does so by using the data from previous samples to generate a statistical model of the system. This model is then used to suggest the next input through an \emph{acquisition function}. The design of the acquisition function is non-trivial; and is critically important to the algorithms performance. It must balance selecting points with the goal of improving the statistical model (\emph{exploration)}, and selecting points with the goal of utilising the improved statistical model to find the global optima \emph{(exploitation)}. In addition to this, the costly nature of these problems means that it is desirable to have a theoretical guarantee of the algorithm's performance. 

There are a large range of acquisition functions, all with different balances of exploration and exploitation, but we focus on \emph{Gaussian process upper confidence bound} (GP-UCB) \cite{srinivas2010} in this work. This controls its exploration-exploitation trade-off with a single hyperparameter, $\beta_t$. It has strong theoretical guarantees on the overall convergence rate, but the bound they give is fairly loose. This causes the value of $\beta_t$ to be too large, causing significant over-exploration and hence poor practical performance. In practice, the theoretical guarantees need to be weakened by selecting a far smaller $\beta_t$. 

We introduce a novel modification to the GP-UCB acquisition function that significantly improves its exploration-exploitation balance while still having a strong convergence guarantee. This is done by sampling its trade-off parameter with a distribution that allows for a range of exploration factors. However, the distribution is chosen such that convergence is guaranteed to be sub-linear while the sampled $\beta_t$ is generally smaller than the traditional GP-UCB. This reduction leads to a direct improvement on the convergence efficiency in practice. We demonstrate this improved performance over the standard GP-UCB implementation in a range of both synthetic benchmark functions and real-world applications.

In summary, our contributions are:
\begin{itemize}
\item The development of a modified acquisition function: RGP-UCB.
\item A convergence analysis of Bayesian optimisation using RGP-UCB.
\item The demonstration of the performance of our method on a range of synthetic and real-world problems.
\end{itemize}
\section{Background}
In this section, we provide a brief overview of Bayesian optimisation, with an emphasis on acquisition functions and regret bounds. For a more in-depth overview of Bayesian optimisation, we refer readers to \cite{Rasmussen2006} and \cite{Brochu2010}.
\subsection{Bayesian Optimisation}
Bayesian optimisation is a method for optimising expensive, noisy black-box functions. It represents the system being optimised as an unknown function, $f$. As this is black-box, it is impossible to directly observe. However, we can sample it with an input variable, $x$, to obtain a noisy output, $y=f(x)+\epsilon_{noise}$, where $\epsilon_{noise}\sim N(0,\sigma_{noise})$ is the random noise corrupting the measurement. Bayesian optimisation seeks to efficiently find the optimal input for such systems over a bounded search space, $\mathcal{X}$:
\begin{equation}
x^*=\underset{x\in\mathcal{X}}{\arg\max}f(x)
\end{equation}
To do this, it creates a statistical model of $f$ using all previously sampled input-output pairs, $D_{t-1}=\{x_i,y_i\}_{i=1}^{t-1}$. This statistical model is usually a \emph{Gaussian process}, but other models can be used. The statistical model is then used to create an \emph{acquisition function}, $\alpha(x)$. This is essentially a map of our belief of how useful a given input will be for optimising the system. As such, it can be used to suggest $x_t$; the next input with which to sample the system. This input and its corresponding output can then be added to the previous data, $D_t=D_{t-1}\cup\{x_t,y_t\}$. This process can be iterated, with the Gaussian process improving with each iteration, until a stopping condition has been met. As Bayesian optimisation is generally used on expensive systems, this stopping condition is often a maximum number of experiments, $T$.
\subsection{Gaussian Process}
The Gaussian process is the most common statistical model used in Bayesian optimisation. It models each point in $x\in\mathcal{X}$ as a Gaussian random variable. As such, it is completely characterised by a mean function, $\mu(x)$, and a variance function, $\sigma^2(x)$. However, in order to model sensibly behaved functions there must be correlation between neighbouring points. This is done by conditioning the distribution on the data via a \emph{kernel function}. There are many viable kernel functions, but one of the simplest and most popular is the \emph{squared exponential kernel} as it only depends on a single hyperparameter, the lengthscale $l$. This is given by
\begin{equation}
k_{SE}(x_{i},x_{j})=\exp\left(-\frac{\left\Vert x_{i}-x_{j}\right\Vert ^{2}}{2l^{2}}\right)
\end{equation}
Using a kernel such as this, a predictive distribution, $f(x)\sim\mathcal{N}(\mu(x)\sigma^2(x))$, can be obtained by conditioning the Gaussian process on the data, $D_t$:
\begin{eqnarray}
\mu_{t}(x)&=&\mathbf{k_{*}}(\mathbf{K}_{t}+\sigma^2_{noise}\mathbf{I})^{-1}\boldsymbol{y},\\\nonumber
\sigma_{t}^{2}(x)&=&k_{*\!*}-\mathbf{k_{*}}(\mathbf{K}_{t}+\sigma^2_{noise}\mathbf{I})^{\mathrm{-1}}\mathbf{k}_{*}^{\mathit{T}}
\end{eqnarray}
where $\mathbf{K}_{t,(i,j)}=k(x_i,x_j)$ is the \emph{kernel matrix},  $k_{*\!*}=k_{t}(x,x)$, and $\mathbf{k}_{*}=[k(x_{1},x),k(x_{2},x),\ldots,k(x_{t},x)]$.
Here $\mathbf{I}$ is the identity matrix with the same dimensions
as $\mathbf{K}_{t}$, and $\sigma_{noise}$ is the output noise standard deviation. 
\subsection{Acquisition Functions}
Once the Gaussian process has been generated, it must then be used to create an \emph{acquisition function}. This is chosen such that its global maxima in $\mathcal{X}$ will be the best next point to sample:
\begin{equation}
x_t=\underset{x\in\mathcal{X}}{\arg\max}\alpha_{t-1}(x)
\end{equation}
However, the design of such a function is non-trivial. It must first suggest points spread over the search space to improve the Gaussian process. This is called \emph{exploration}. Once the Gaussian process has been improved enough, it must then transition to suggesting points in regions that have a high probability of containing the global optima. This is called \emph{exploitation}. If an acquisition function does not explore enough, it may get stuck exploiting sub-optimal regions and never find the global optima. However, if it explores too much, it may waste costly evaluations improving an already adequate Gaussian process. This makes balancing exploration and exploitation vital.
There is a wide range of common acquisition functions, all of which have different balances of exploration and exploitation. These include entropy search (ES) by \cite{hennig2012entropy}, predictive entropy search (PES) by \cite{hernandez2014predictive}, knowledge gradient (KG) by \cite{scott2011correlated}, and others. However, we will only consider the \emph{Gaussian process upper confidence bound} (GP-UCB) by \cite{srinivas2010}, Thompson sampling by \cite{russo2014learning}, and \emph{expected improvement} (EI) by \cite{jones1998efficient}, with the latter two only being used as baselines.
%\subsubsection{Regret}
%Bayesian optimisation is commonly used in high value problems. As such, theoretical measures of its convergence are desirable. The \emph{cumulative regret} is one such measure \cite{srinivas2010}. Regret is simply the difference between the current sampled value and the global optima. The cumulative regret is the sum of the regret over all iterations:
%\begin{equation}
%R_T=\sum_{t=1}^T\left[f(x^*)-f(x_t)\right]
%\end{equation}
%This was extended by \cite{russo2014learning} to the case where $x_t$ are chosen in a probabilistic rather than deterministic manner. They introduced the idea of \emph{Bayesain regret}:
%\begin{equation}
%BR_T(T)=\sum_{t=1}^T\mathbb{E}\left[f(x^*)-f(x_t)\right]
%\end{equation}
\subsubsection{GP-UCB}
GP-UCB \cite{srinivas2010} is one of the most intuitive acquisition functions. It balances exploration and exploitation through a single hyperparameter, $\beta_t$:
\begin{equation}
\alpha_t^{GP-UCB}(x)=\mu_t(x)+\sqrt{\beta_t}\sigma_t(x)
\end{equation}
Increasing $\beta_t$ makes the acquisition function favour points with high variance, causing more exploration. Decreasing $\beta_t$ will make the acquisition function favour points with high mean, causing more exploitation. However, the selection of $\beta_t$ is not done to optimally balance exploitation and exploration, but is done such that the cumulative regret is bounded. It has been proved that, assuming the chosen kernel satisfies
\begin{equation}
P\left\lbrace\sup_{x\in \mathcal{X}}|\partial f/\partial x_i|>L\right\rbrace\leq ae^{-(L/b)^2},\quad i=1\ldots t
\end{equation}
for some constants $a,b>0$, then with probability $1-\delta$, the algorithm will have sub-linear regret if
\begin{equation}
\beta_t=2\log(t^{2}\pi^2/(3\delta))+2d\log\left(t^2dbr\sqrt{\log(4da/\delta)}\right)
\end{equation}
While the regret bound provided by this choice of $\beta_t$ is desirable, it unfortunately is far larger than needed. This leads to sub-optimal real world performance due to over-exploration. In their own paper, the authors divided the suggested $\beta_t$ by a factor of 5 to achieve better performance \cite{srinivas2010}.
\section{Proposed Method}
In this section we describe our improved GP-UCB acquisition function, \emph{randomised Gaussian process upper confidence bound} (RGP-UCB), and prove that it has a sub-linear regret bound.
\subsection{RGP-UCB}
\begin{figure}
\centering
\includegraphics[width=0.8\columnwidth]{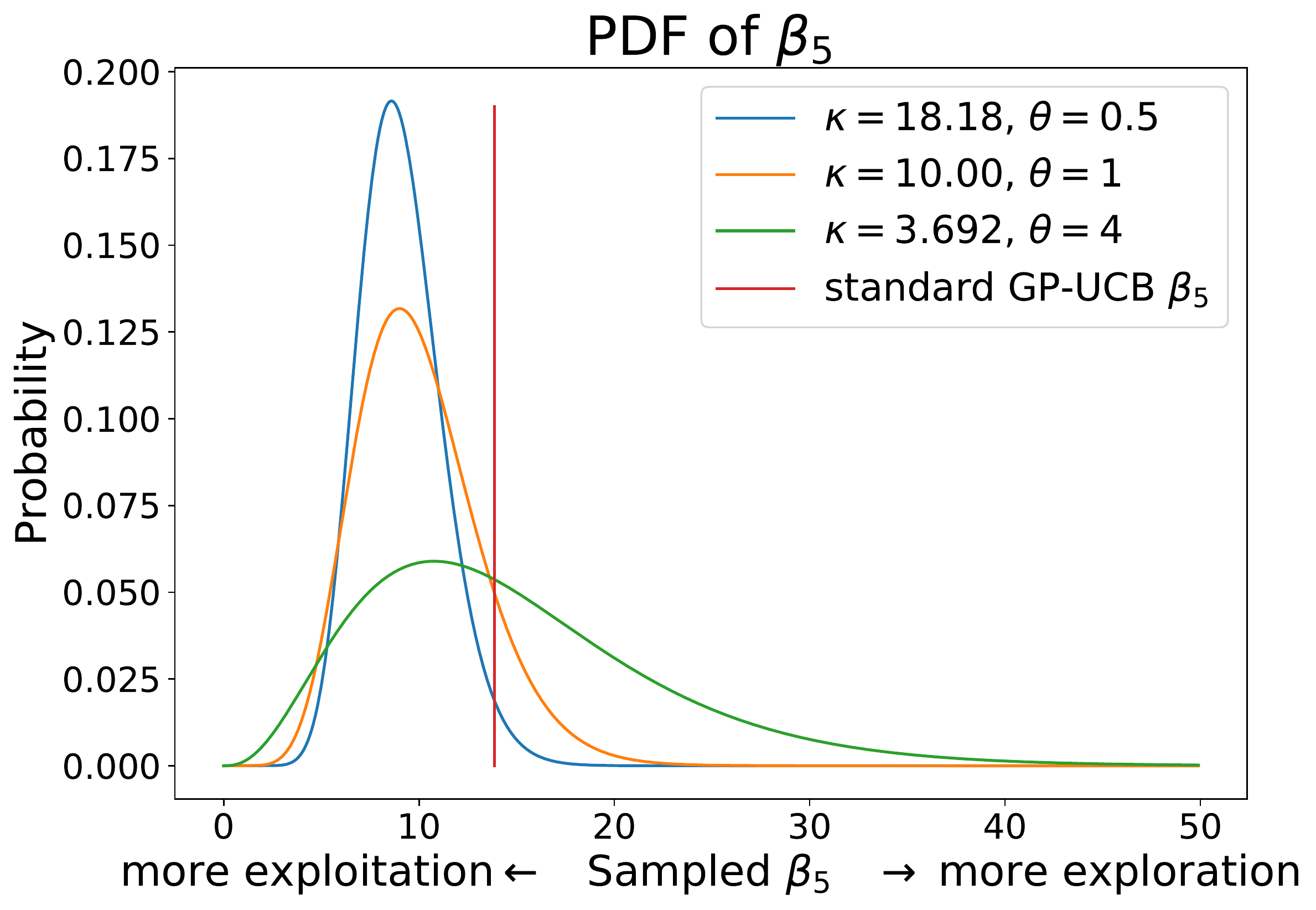}

\caption{A range of $\beta_5$ distributions with $t=5$. The parameters are chose to satisfy \textbf{Theorem \ref{trm:GammaParams}}. Note that increasing $\theta$ shifts the distribution right, \emph{increasing exploration}. \label{fig:Beta}}
\end{figure}

While the standard GP-UCB method has a desirable regret bound, it has relatively poor performance. This is due to the $\beta_t$ used to satisfy this bound being far too large, forcing significant over-exploration. As such, a method for selecting a smaller $\beta_t$ while maintaining a regret bound is desirable. We show that this can be done by sampling $\beta_t$ from a distribution, and as such, we call our method \emph{randomised Gaussian process upper confidence bound} (RGP-UCB). Doing so means that we bound the Bayesian regret instead of the regret, but it allows for far greater freedom in selecting $\beta_t$, letting it be set far smaller while still maintaining convergence guarantees. However, we require $\beta_t>0$ since a negative $\beta_t$ will punish exploration. We also do not want our distribution to suggest a very large $\beta_t$ as that will cause over-exploration. As such, we draw $\beta_t$ from a $\Gamma(\kappa_t,\theta)$ distribution. Examples of this distribution can be seen in \textbf{Figure \ref{fig:Beta}} and the complete algorithm is shown in \textbf{Algorithm \ref{alg:RGP_UCB_alg}}. Much like with standard GP-UCB, the parameters of this distribution are chosen to satisfy a regret bound, as per \textbf{Theorem \ref{trm:GammaParams}}. However, we show that we only need to set one of the two distribution parameters for the bound to hold. Unlike the standard GP-UCB, this allows us to tune $\beta_t$ to substantially improve our methods performance without compromising its theoretical guarantees.

\begin{algorithm}[tb]
\caption{Bayesian Optimisation with RGP-UCB}
\label{alg:RGP_UCB_alg}
\textbf{Input}:$D_{t_0}=\lbrace x_i,y_i\rbrace_{i=1}^{t_0}$, $\Gamma$ scale parameter, $\theta$, \# Iterations $T$, Kernel lengthscale $l$
\begin{algorithmic}[1] %[1] enables line numbers
\FOR{$t=t_0$ to $T$}
\STATE Build a Gaussian Process (GP) with $D_{t}$.
\STATE Set $\kappa_t=\frac{\log\left(\frac{1}{\sqrt{2\pi}}{(t^2+1)}\right)}{\log\left(1+\theta/2\right)}$ and draw $\beta_t$ from $\Gamma(\kappa_t,\theta)$
\STATE Optimise the acquisition function, $\alpha_t(x)=\mu_t(x)+\sqrt{\beta_t}\sigma_t(x)$ to obtain $x_{t+1}$ and use it to sample $y_{t+1}=f(x_{t+1})+\epsilon_n$ from the system.
\STATE Augment the data, $D_{t+1}=D_y\cup\{x_{t+1},y_{t+1}\}$.
\ENDFOR
\RETURN $x^+=x_i$ such that $y_i=\underset{t\in [1,T]}{\max} y_t$.

\end{algorithmic}
\end{algorithm}

\subsection{Theoretical Analysis}

Bayesian optimisation is commonly used in high value problems. As such, theoretical measures of its convergence are desirable. The \emph{cumulative regret} is one such measure, and is the cornerstone of GP-UCB \cite{srinivas2010}. Regret is simply the difference between the current sampled value and the global optima. The cumulative regret is the sum of the regret over all iterations:
\begin{equation}
R_T=\sum_{t=1}^T\left[f(x^*)-f(x_t)\right]
\end{equation}
As RGP-UCB is probabilistic, we instead need to use the \emph{Bayesain regret} by Russo et al \cite{russo2014learning}:
\begin{equation}
BR_T=\sum_{t=1}^T\mathbb{E}\left[f(x^*)-f(x_t)\right]
\end{equation}

However, their proof was for Gaussian processes with a finite search space, i.e. $|\mathcal{X}|< \infty$. As such, we follow a method similar to \cite{srinivas2010} and \cite{kandasamy2017asynchronous} and introduce a discretisation of our search space into an $\tau^d$ grid of equally spaced points, $\mathcal{X}_{\mathsf{dis}}$. We denote $[x]_{\tau}$ as the closest point to $x$ in $\mathcal{X}_{\mathsf{dis}}$.

With this, we can begin bounding the Bayesian regret of our algorithm by decomposing it into components that are easier to bound.

\begin{lem}
The Bayesian regret of a probabilistic RGP-UCB algorithm, $\alpha_{t-1}$, over $T$ iterations can be decomposed as
\begin{eqnarray}
BR_T&\leq&\underbrace{\sum_{t=1}^T\mathbb{E}\left[\alpha_{t-1}([x_t]_{\tau})-f([x_t]_{\tau})\right]}_{R_1}\nonumber\\
&&+\underbrace{\sum_{t=1}^T\mathbb{E}\left[f([x^*]_{\tau})-\alpha_{t-1}([x^*]_{\tau})\right]}_{R_2}\nonumber\\
&&+\underbrace{\sum_{t=1}^T\mathbb{E}\left[f(x^*)-f([x^*]_{\tau})\right]}_{R_3}\nonumber\\
&&+\underbrace{\sum_{t=1}^T\mathbb{E}\left[f([x_t]_{\tau})-f(x_t)\right]}_{R_4}
\end{eqnarray}
\end{lem}

\begin{proof}
%By the definition, of $x_t=\max_{x\in\mathcal{X}}\alpha_{t-1}(x)$, we have that $\alpha(x_t)\geq \alpha(x^*)$. 
This simply follows from the fact that, as $x_t=\arg\max_{x\in\mathcal{X}}\alpha_{t-1}(x)$, we have that $\alpha(x_t)\geq \alpha(x^*)$.
%\begin{eqnarray}
%BR_T&=&\sum_{t=1}^T\mathbb{E}[f(x^*)-f(x_t)]\nonumber\\
%%&=&\sum_{t=1}^T\mathbb{E}[f(x^*)-f(x_t))\nonumber\\
%%&&+\alpha_{t-1}([x^*]_{\tau})-\alpha_{t-1}([x^*]_{\tau})\nonumber\\
%%&&+f([x^*]_{\tau})-f([x^*]_{\tau})\nonumber\\
%&\leq&f([x_t]_{\tau})-f([x_t]_{\tau})]\nonumber\\
%&&+\sum_{t=1}^T\mathbb{E}[\alpha_{t-1}([x_t]_{\tau})-f([x_t]_{\tau})]\nonumber\\
%&&+\sum_{t=1}^T\mathbb{E}[f([x^*]_{\tau})-\alpha_{t-1}([x^*]_{\tau})]\nonumber\\
%&&+\sum_{t=1}^T\mathbb{E}[f([x_t]_{\tau})-f(x_t))]\nonumber\\
%&&+\sum_{t=1}^T\mathbb{E}[f(x^*)-f([x^*]_{\tau})]
%\end{eqnarray}
\end{proof}
With this decomposition, we simply need to find a bound for each term to bound the Bayesian regret. We will start with the second term.
\begin{theorem}
Assuming that $\beta_t$ is drawn from a $\Gamma(\kappa_t,\theta)$ distribution with $\kappa_t=\frac{\log\left(\frac{1}{\sqrt{2\pi}}{(t^2+1)}\right)}{\log\left(1+\theta/2\right)}$, the following bound holds
\begin{eqnarray}\label{eqn:R2}
R_2&=&\sum_{t=1}^T\mathbb{E}\left[f([x^*]_{\tau})-\alpha_{t-1}([x^*]_{\tau})\right] \nonumber\\
&\leq& \sum_{t=1}^{T}\frac{1}{t^2+1}
\end{eqnarray}
\end{theorem}
\begin{proof}
As the posterior distribution of $f(x)$ at iteration $t-1$ is $\mathcal{N}(\mu_{t-1},\sigma^2_{t-1})$, the distribution of $f(x)-\alpha_{t-1}(x)|\beta_t$, is simply $\mathcal{N}(-\sqrt{\beta_{t}}\sigma_{t-1}(x)$,$\sigma_{t-1}^2(x)$). Hence
\begin{eqnarray}
\mathbb{E}_{\beta}&&\hspace{-2.5em}\left[\mathbb{E}_f\left[\mathbf{1}\left\lbrace f(x)-\alpha_{t-1}(x)\geq 0\right\rbrace [f(x)-\alpha_{t-1}(x)|\beta_{t}]\right]\right]\nonumber\\
&\leq&\mathbb{E}_{\beta}\left[\frac{\sigma_{t-1}(x)}{\sqrt{2\pi}}\exp\left\lbrace-\frac{\sqrt{\beta_{t}}}{2}\right\rbrace\right]\nonumber\\
&\leq& \frac{\sigma_{t-1}(x)}{\sqrt{2\pi}}\mathbb{E}_{\beta}\left[\exp\left\lbrace-\frac{\sqrt{\beta_{t}}}{2}\right\rbrace\right]
\end{eqnarray}
We note that the exponential term is simply the moment generating function of $\sqrt{\beta_{t}}$, which has a closed form for a gamma distribution. This lets us express our inequality as

\begin{eqnarray}
\mathbb{E}_{\beta}&&\hspace{-1cm}\left[\mathbb{E}_f\left[\mathbf{1}\left\lbrace f(x)-\alpha_{t-1}(x)\geq 0\right\rbrace [f(x)-\alpha_{t-1}(x)|\beta_{t}]\right]\right]\nonumber\\
&\leq& \frac{\sigma_{t-1}(x)}{\sqrt{2\pi}}\frac{1}{(1+\theta_{t-1}/2)^{\kappa_{t-1}}}
\end{eqnarray}
As we want this to decay at a sub-linear rate, we need
\begin{eqnarray}\label{eqn:k}
&&\frac{\sigma_{t-1}(x)}{\sqrt{2\pi}}\frac{1}{(1+\theta_{t-1}/2)^{\kappa_{t-1}}}
\leq \frac{1}{(t^2+1)|\mathcal{X}_{\mathsf{dis}}|}\nonumber\\
&&\kappa_{t-1}\geq  \frac{\log\left(\frac{1}{\sqrt{2\pi}}{\sigma_{t-1}(x)(t^2+1)|\mathcal{X}_{\mathsf{dis}}|}\right)}{\log\left(1+\theta_{t-1}/2\right)}
\end{eqnarray}
Setting $\kappa_{t-1}=\frac{\log\left(\frac{1}{\sqrt{2\pi}}{(t^2+1)}\right)}{\log\left(1+\theta_{t-1}/2\right)}$ to satisfy this, equation $R_2$ is bounded by the following:
\begin{eqnarray}
R_2&=&\sum_{t=1}^T\mathbb{E}\left[f([x^*]_{\tau})-\alpha_{t-1}([x^*]_{\tau})\right]\nonumber\\
&\leq&\sum_{t=1}^{T}\sum_{x\in \mathcal{X}}\mathbb{E}_{\beta_{t}}[\mathbb{E}_f[\mathbf{1}\lbrace f(x)\nonumber\\
&-&\alpha_{t-1}(x)\geq 0\rbrace [f(x)-\alpha_{t-1}(x)|\beta_{t}]]]\nonumber\\
&\leq&\sum_{t=1}^{T}\frac{1}{t^2+1}
\end{eqnarray}
\end{proof}
Next, we attempt to bound the first component.
\begin{theorem}
Assuming that $\beta_t$ is drawn from a $\Gamma(k,\theta)$ distribution, the following bound holds
\begin{eqnarray}
R_1&=&\sum_{t=1}^T\mathbb{E}\left[\alpha_{t-1}([x_t]_{\tau})-f([x_t]_{\tau})\right]\nonumber\\
&\leq&\sqrt{\left[1+\frac{k-1}{F^{-1}\left(1-\frac{1}{T}\right)}\right]\gamma+F^{-1}\left(1-\frac{1}{T}\right)}\nonumber\\&&\times\sqrt{T\sum_{t=1}^T\sigma_{t-1}(x_t)}
\end{eqnarray}
\end{theorem}
\begin{proof}
Using Jensen's Inequality we have
\begin{eqnarray}
R_1&=&\sum_{t=1}^T\mathbb{E}\left[\alpha_{t-1}([x_t]_{\tau})-f([x_t]_{\tau})\right]\nonumber\\
&=&\sum_{t=1}^T\mathbb{E}_{\beta_{t}}\left[\mathbb{E}_f\left[\alpha_{t-1}([x_t]_{\tau})-f([x_t]_{\tau})|\beta_{t}\right]\right]\nonumber\\
&=&\sum_{t=1}^T\mathbb{E}_{\beta_{t}}\left[\sqrt{\beta_{t}}\sigma_{t-1}([x_t]_{\tau})\right]\nonumber\\
&\leq&\sum_{t=1}^T\sigma_{t-1}([x_t]_{\tau})\sqrt{\mathbb{E}_{\beta_{t}}\left[\beta_{t}\right]}
\end{eqnarray}
We can then use the Cauchy-Schwartz inequality to get
\begin{eqnarray}
R_1&\leq&\sqrt{\sum_{t=1}^T\mathbb{E}_{\beta_{t}}\left[\beta_{t}\right]}\sqrt{\sum_{t=1}^T\sigma^2_{t-1}([x_t]_{\tau})}\nonumber\\
&\leq&\sqrt{T\mathbb{E}_{\beta_{t}}\left[\underset{t\leq T}{\max}\beta_{t}\right]}\sqrt{\sum_{t=1}^T\sigma^2_{t-1}([x_t]_{\tau})}
\end{eqnarray}
As $\beta_{t}$ is a gamma distribution with shape parameter $\kappa_t$ and scale parameter $\theta$, its maximum is given by
\begin{equation}
\mathbb{E}\left[\underset{t\leq T}{\max}\beta_t\right]\approx\left[1+\frac{\kappa_t-1}{F^{-1}\left(1-\frac{1}{T}\right)}\right]\gamma+F^{-1}\left(1-\frac{1}{T}\right)
\end{equation}
where $\gamma$ is the Euler-Mascheroni constant and $F^{-1}(x)$ is the inverse CDF of $\beta$. 
This finally gives us the following bound:
\begin{eqnarray}
R_i&\leq&\sqrt{\left[1+\frac{k-1}{F^{-1}\left(1-\frac{1}{T}\right)}\right]\gamma+F^{-1}\left(1-\frac{1}{T}\right)}\nonumber\\&&\times\sqrt{T\sum_{t=1}^T\sigma^2_{t-1}(x_t)}
\end{eqnarray}
\end{proof}
Finally, we need to bound components $R_3$ and $R_4$. For these, we can use lemma 10 from \cite{kandasamy2017asynchronous}.
\begin{lem}
At step $t$, for all $x\in\mathcal{X}$, $\mathbb{E}[|f(x)-f([x]_{\tau})|]\leq\frac{1}{2t^2}$.
\end{lem}
This means that we have that
\begin{eqnarray}
R_3+R_4&\leq&2\sum^{T}_{t=1}\frac{1}{2t^2}\nonumber\\
&\leq&\frac{\pi^2}{6}
\end{eqnarray}
With this we can finally find our Bayesian regret bound.
\begin{theorem}\label{trm:GammaParams}
If $\beta_t$ is sampled from a $\Gamma(\kappa_t,\theta)$ distribution with
\begin{equation}
\kappa_t=\frac{\log\left(\frac{1}{\sqrt{2\pi}}{(t^2+1)}\right)}{\log\left(1+\theta/2\right)}
\end{equation}
then the RGP-UCB acquisition function has its Bayesian regret bounded by
\begin{eqnarray}
BR_T&\leq&\sqrt{\left[1+\frac{\kappa_t-1}{F^{-1}\left(1-\frac{1}{T}\right)}\right]\gamma+F^{-1}\left(1-\frac{1}{T}\right)}\nonumber\\&&\times\sqrt{T\sum_{t=1}^T\sigma^2_{t-1}(x_t)}+\sum_{t=1}^{T}\frac{1}{t^2+1}+\frac{\pi^2}{6}
\end{eqnarray}
where $\gamma$ is the Euler-Mascheroni constant and $F^{-1}(x)$ is the inverse CDF of $\beta_t$
\end{theorem}
\begin{proof}
The result follows simply by combining the bounds for the various components.
\end{proof}

\section{Results}
In this section we present results that demonstrate the performance of RGP-UCB in comparison to other common acquisition functions. We also demonstrate the impact of varying the $\theta$ parameter of the gamma distribution used to sample $\beta_t$. The Python code used for this paper can be found at \textbf{ \url{https://github.com/jmaberk/RGPUCB}}.
\subsection{Experimental Setup}
We test our method against a selection of common acquisition functions on a range of Bayesian optimisations problems. These include a range of synthetic benchmark functions and real-world optimisation problems. In each case, the experiment was run for $40d$ iterations and repeated 10 times with $3d+1$ different initial points. The initial points are chosen randomly with a Latin hypercube sample scheme \cite{jones2001taxonomy}. The methods being tested are:
\begin{itemize}
\item Our randomised Gaussian process upper confidence bound with $\theta=8$ (RGP-UCB $\theta=8$).
\item Our randomised Gaussian process upper confidence bound with $\theta=1$ (RGP-UCB $\theta=1$).
\item Our randomised Gaussian process upper confidence bound with $\theta=0.5$ (RGP-UCB $\theta=0.5$).
\item Standard Gaussian process upper confidence bound (GP-UCB) \cite{srinivas2010}.
\item Expected improvement (EI) \cite{jones1998efficient}.
\item Thompson sampling (Thompson) \cite{russo2014learning}.
\end{itemize}
Note that we turn all functions that are traditionally minimised into maximisation problems by taking their negative for consistency. As such, higher results are always better.
\subsection{Selection of the Trade-off Parameter}
An advantage of our method is that it can change its exploration-exploitation balance without compromising its convergence guarantee. This is done by changing the $\theta$ parameter in the $\beta_t\sim\Gamma(\kappa_t,\theta)$ distribution. Increasing $\theta$ will increase the expected $\beta_t$, increasing exploration.

As different problems favour different exploration-exploration balances, we tested a range of $\theta$ values on a range of different problems. In \textbf{Figure \ref{fig:trade-off}}, we show the performance of a range of $\theta$ values on an exploitation favouring problem, the Alpine 2 (5D) function, and an exploration favouring problem, the Dropwave (2D) function.

\begin{figure}
\centering
\begin{tabular}{|l |c| c|}
\hline
&Dropwave (2D) & Alpine 2 (5D)  \\ \hline
$\theta=0.1$& 0.738$\pm$ 6.7e-2 &78.9$\pm$12.4 \\ \hline
$\theta=0.5$& 0.755$\pm$ 5.0e-2 &\textbf{92.1$\pm$12.2}  \\ \hline 
$\theta=1$& 0.754$\pm$ 6.9e-2 &77.8$\pm$12.7  \\ \hline
$\theta=2$& 0.727$\pm$ 8.6e-2 &77.5$\pm$12.5  \\ \hline
$\theta=4$& 0.847$\pm$ 5.7e-2 &71.5$\pm$13.6  \\ \hline
$\theta=8$& \textbf{0.848$\pm$ 3.3e-2} &43.4$\pm$9.84  \\ \hline
$\theta=16$& 0.814$\pm$ 6.2e-2 &45.4$\pm$10.0  \\ \hline
\end{tabular}
\caption{Best found values using different $\theta$ parameters on an exploitation favouring function (Alpine 2) and an exploration favouring function (Dropwave).\label{fig:trade-off}}
\end{figure}
It was found that $\theta=8$ gives good performance on both the above exploration-favouring problem and other similar problems tested. Likewise, $\theta=0.5$ is a good choice for exploitation favouring problems. We also note that $\theta=1$ has decent performance on both problems, making it a good choice for problems where the required exploration-exploitation balance is completely unknown.

\subsection{Synthetic Benchmark Functions}
\begin{figure*}
\centering
\includegraphics[height=4.2cm]{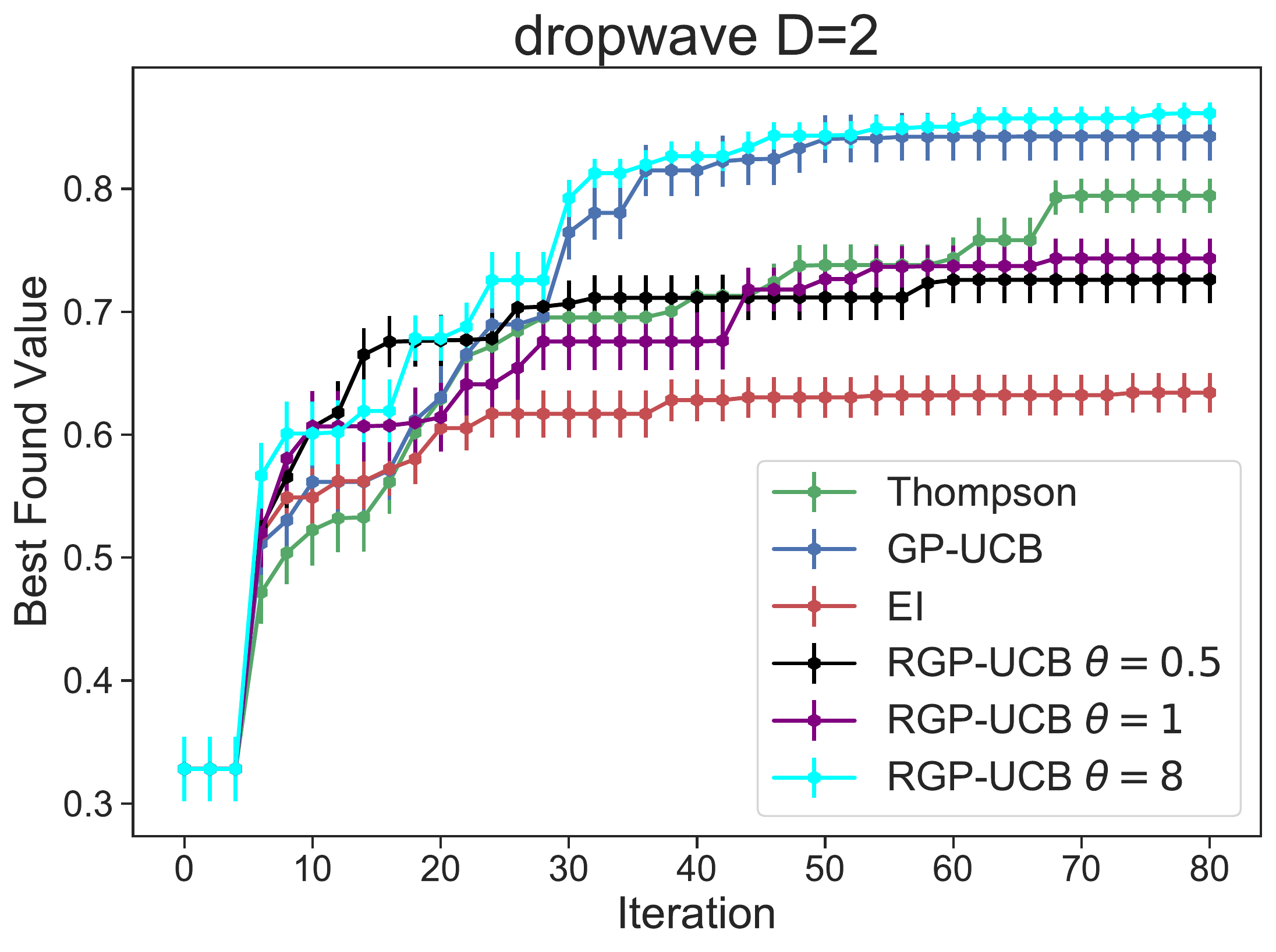}
\includegraphics[height=4.2cm]{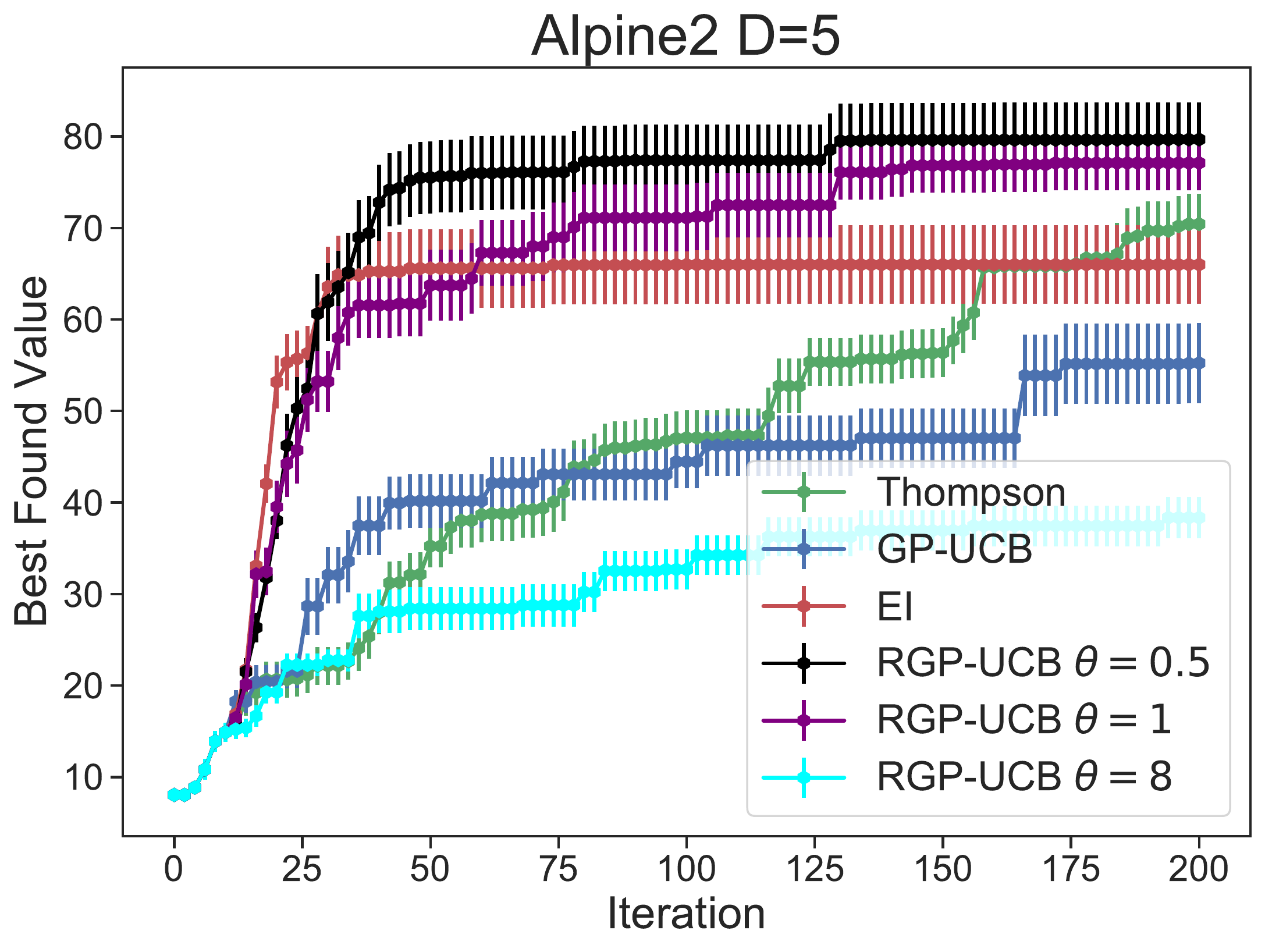}
\includegraphics[height=4.2cm]{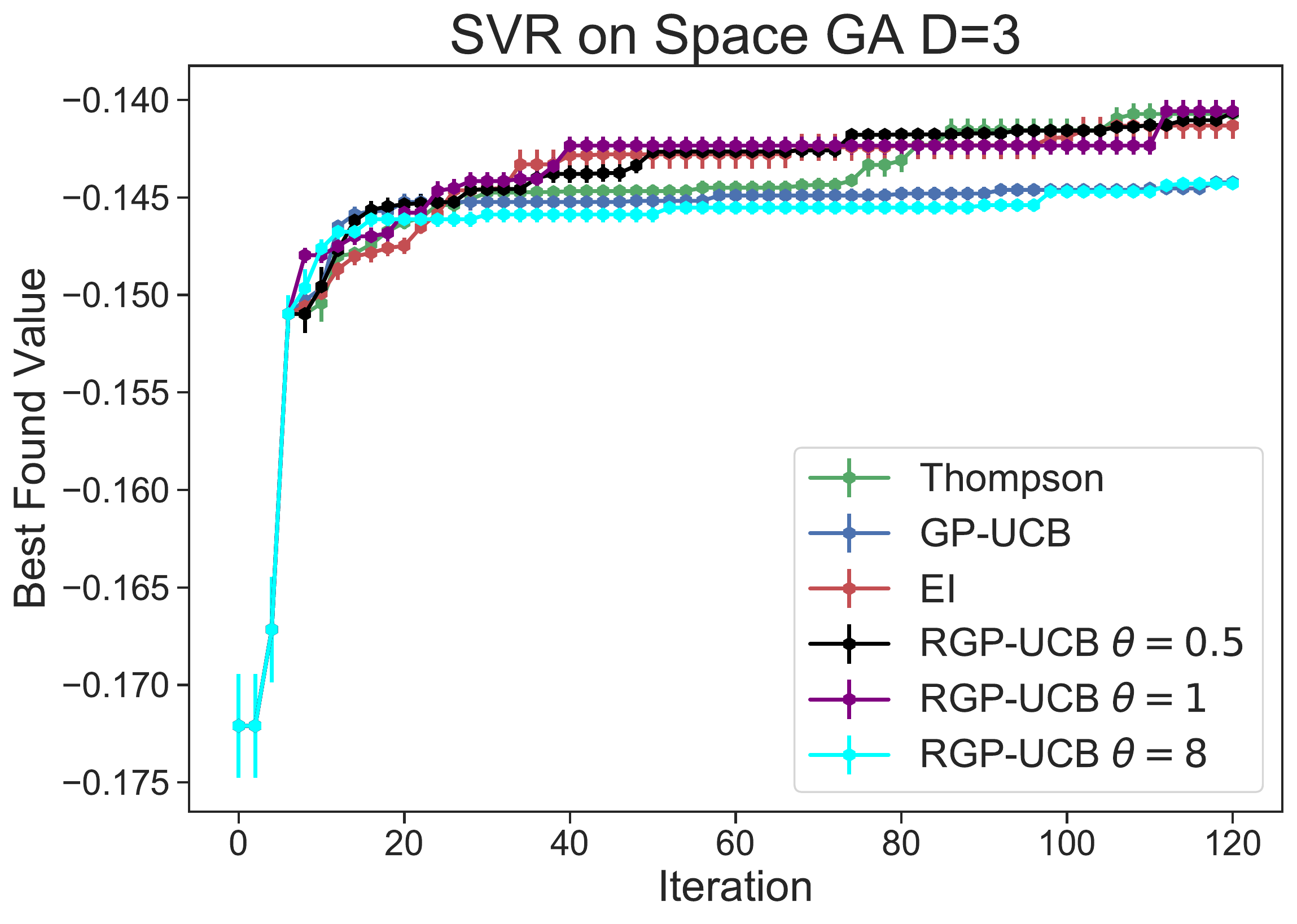}
\includegraphics[height=4.2cm]{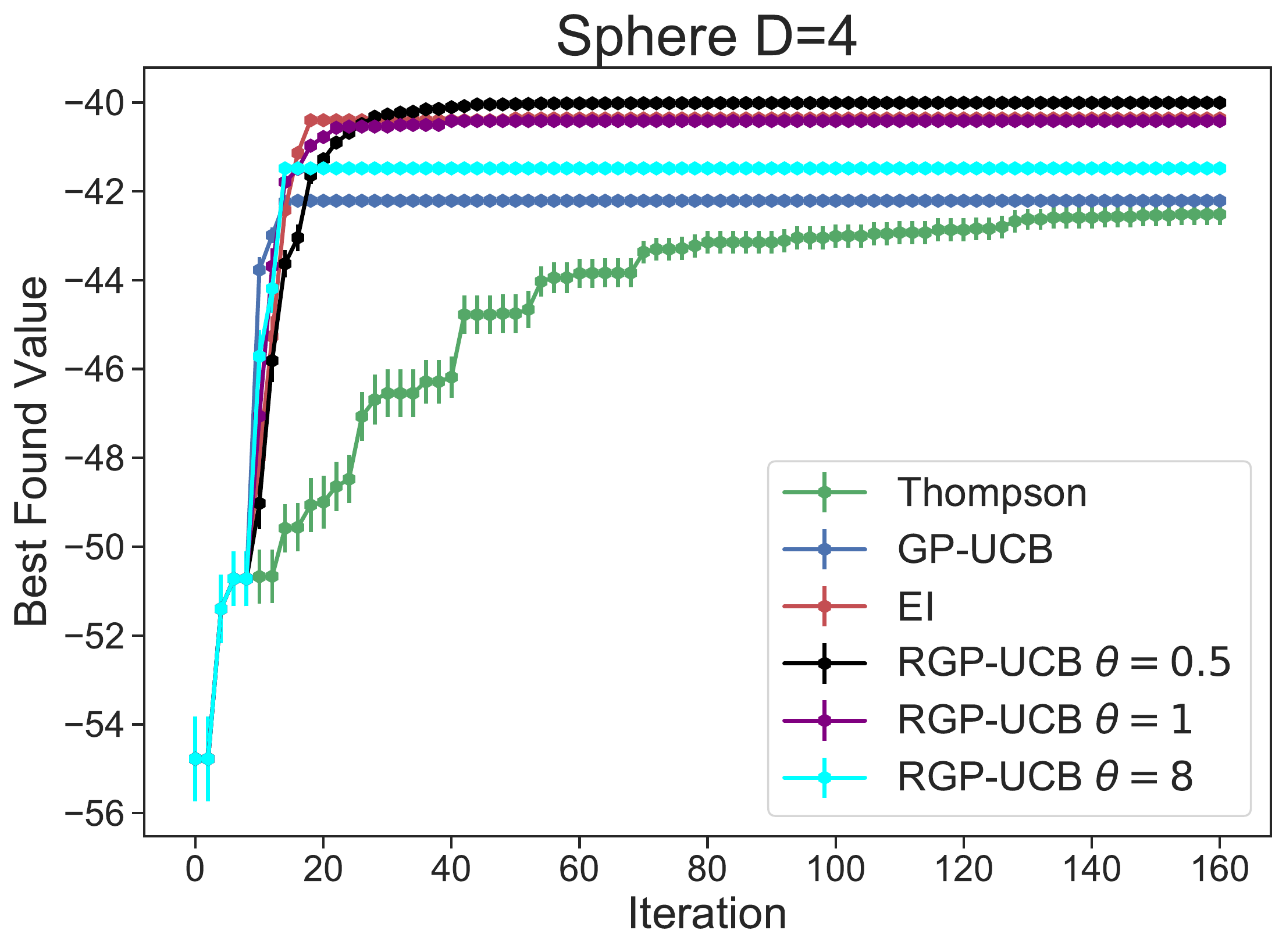}
\includegraphics[height=4.2cm]{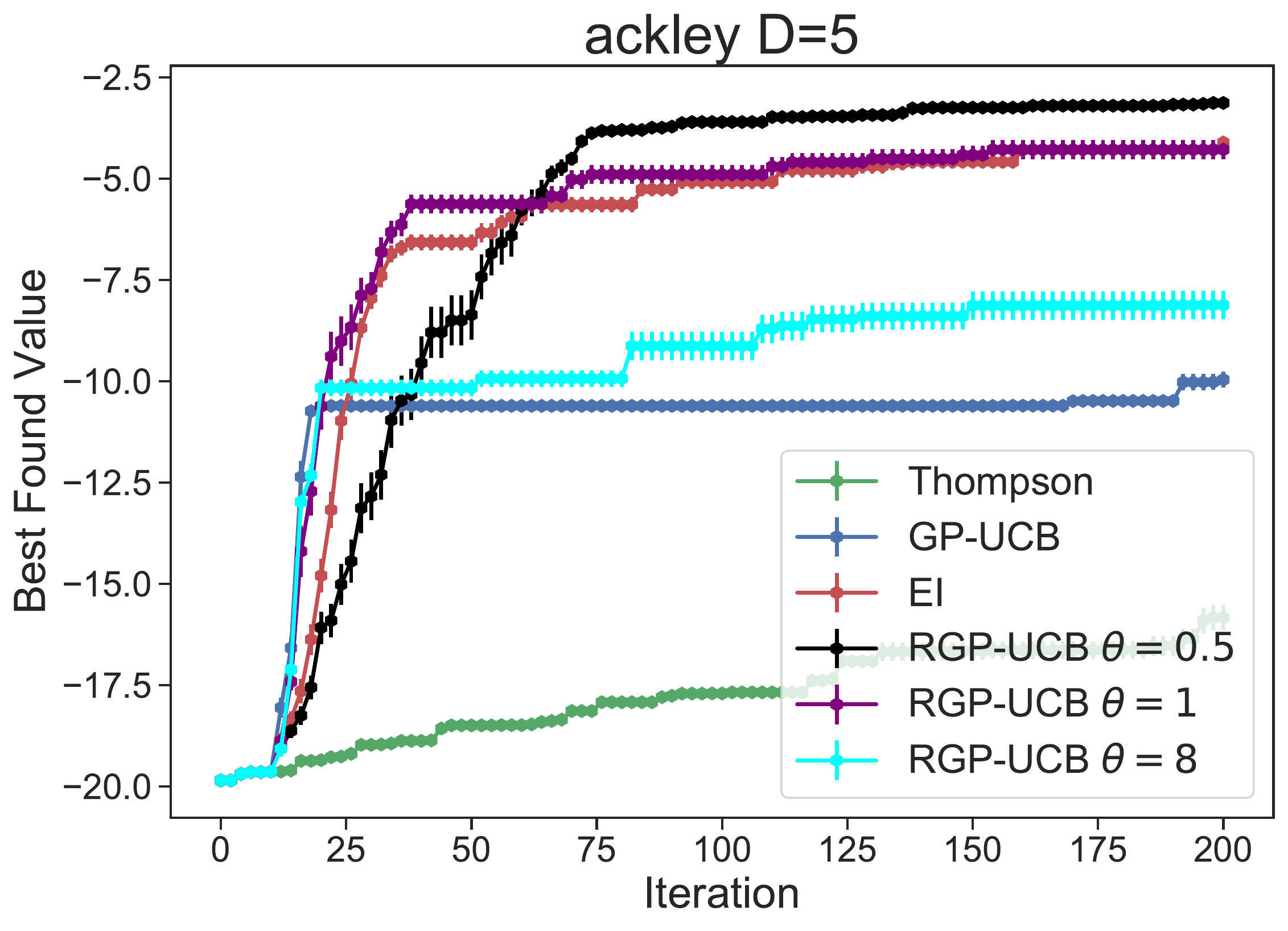}
\includegraphics[height=4.2cm]{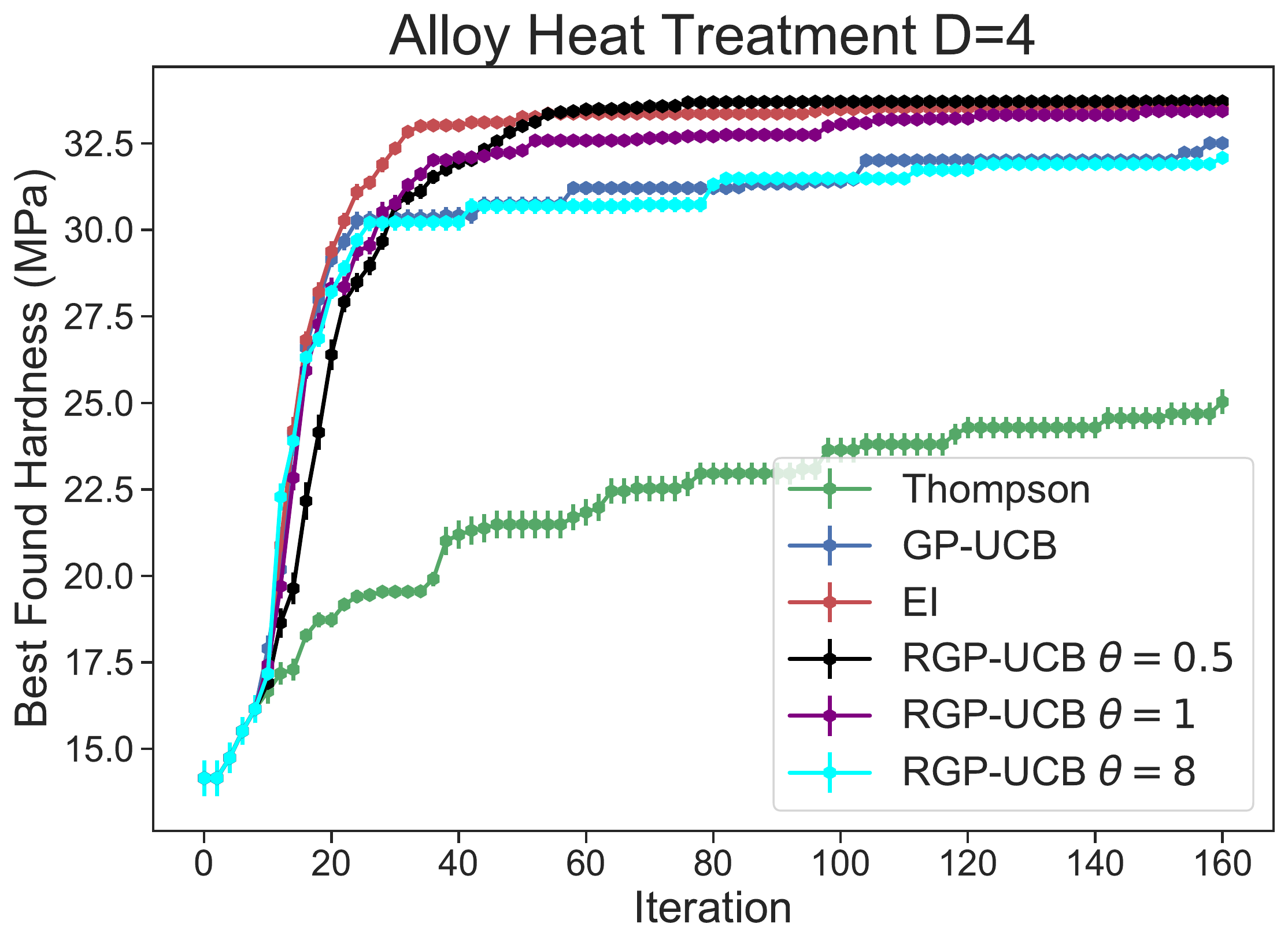}

\caption{Performance of the RGP-UCB acquisition function compared to other common methods on a range of  optimisation problems, including synthetic (left and middle) and real-world (right) examples. \label{fig:synthetic}}
\end{figure*}

The first demonstration of our methods performance is the optimisation of several common synthetic benchmark functions. These are the Dropwave (2D), Sphere (4D), Alpine 2 (5D), and Ackley (5D) functions\footnote{All benchmark functions use the recommended parameters from \url{https://www.sfu.ca/~ssurjano/optimization.html}}. Results for these are shown in\textbf{ Figure \ref{fig:synthetic}}. 
%\begin{figure*}
%\centering
%\begin{tabular}{|c |c| c| c| c| c| c| c|}
%\hline
% & RGP-UCB $\theta=0.5$ & RGP-UCB $\theta=1$ & RGP-UCB $\theta=8$ & EI & GP-UCB & Thompson \\ \hline
%Dropwave (2D) & 0 &0 & 0 &0 & 0 & 0 \\\hline
%Sphere (4D) & 0 &0 & 0 &0 & 0 & 0 \\\hline
%Alpine 2 (5D) & 0 &0 & 0 &0 & 0 & 0 \\\hline
%Alpine 1 (5D) & 0 &0 & 0 &0 & 0 & 0 \\\hline
%Ackley (5D) & 0 &0 & 0 &0 & 0 & 0 \\\hline
%Levy (5D) & 0 &0 & 0 &0 & 0 & 0 \\\hline
%\end{tabular}
%\caption{Performance of the RGP-UCB acquisition function compared to other common methods on a range of synthetic optimisation problems. \label{fig:synthetic}}
%\end{figure*}
Here we can see that RGP-UCB has competitive performance in all of the above cases. In general, it does significantly better than the standard GP-UCB and the Thompson sampling acquisition functions. EI has better early performance in many cases, as it starts exploiting earlier. However, RGP-UCB tends to have better exploration and therefore often able to beat it in the long-term.

The leftmost functions were chosen to be pathological cases which disproportionately favours exploration (Dropwave 2D) and exploitation (Sphere 4D). These can be seen as best-case examples for GP-UCB and EI respectively. However, RGP-UCB is able to out-perform them even on these if its $\theta$ parameter is chosen properly, and does so while maintaining its convergence guarantee. This formulation of EI does not have a known regret bound as it follows the standard implementation and hence doesn't satisfy the assumptions required by the current bounds \cite{bull2011convergence,ryzhov2016convergence,wang2014theoretical,nguyen2017regret}. %It has noise, uses the standard incumbent of $y^+$, and uses the standard iteration budget stopping condition.

In the middle two plots, RGP-UCB is superior even with the conservative parametrisation of $\theta=1$.

\subsection{Machine Learning Hyperparameter Tuning}

Our first demonstration of the real-world performance of RGP-UCB is the hyperparameter tuning of a support vector regression (SVR) \cite{drucker1997support} algorithm. This is the support vector machine classification algorithm extended to work on regression problems, with performance measured in root mean squared error (RMSE). It has three hyperparameters, the threshold, $\epsilon$, the kernel parameter, $\gamma$, and a soft margin parameter, $C$. All experiments are done with the public Space GA scale dataset \footnote{ Dataset can be found at \url{https://www.csie.ntu.edu.tw~cjlin/libsvmtools/datasets/regression.html}}. The results are shown in\textbf{ Figure \ref{fig:synthetic}}.

We can see that the final performance of all three variants of our method exceeds that of standard GP-UCB. The high exploitation and balanced variants are competitive with EI, with the former achieving higher final performance. As with many real-world problems, SVR is known to favour higher exploitation, and is therefore an example of when the user would know to try a smaller $\theta$.

\subsection{Materials Science Application: Alloy Heat Treatment}
The second demonstration of RGP-UCB's performance is the optimisation of a Aluminium-Scandium alloy heat treatment simulation \cite{robson2003extension}. The goal of the simulation is to optimise the resulting alloys hardness, measured in MPa. The hardening process is controlled through multiple cooking stages, each with two hyperparametrs, the duration and a temperature. As we use a two-stage cooking simulation, there is a total of four hyperparameters to optimise through Bayesian optimisation. The results are shown in \textbf{Figure \ref{fig:synthetic}}. 

%\begin{figure}
%\centering
%\includegraphics[width=0.49\columnwidth]{figs/"SVR on Space GA_3_E3I".pdf}
%\includegraphics[width=0.46\columnwidth]{figs/AlloyCooking_Profiling_4_E3I.pdf}
%\caption{Performance of the GP-UCB methods compared against that of other common methods on tuning the hyperparameters for a SVR algorithm (left) and a two step Aluminum-Scandium alloy cooking simulation (right). \label{fig:real_exp}}
%\end{figure}

The results are very similar to the previous SVR example, with the high-exploitation method having the best performance and the balance method being competitive with EI.

\subsection{Conclusion}
We have developed a modified UCB based acquisition function that has substantially improved performance while maintaining a sub-linear regret bound. We have proved that this bound holds in terms of Bayesian regret while allowing for some flexibility in the selection of its parameters. We have also demonstrated the impact of said parameters on the performance. Moreover, we have shown that its performance is competitive or greater than existing methods in a range of synthetic and real-world applications.

\subsubsection*{Acknowledgements}
This research was supported by an Australian Government Research Training Program
(RTP) Scholarship awarded to JMA Berk, and was partially funded by the Australian
Government through the Australian Research Council (ARC). Prof Venkatesh is the
recipient of an ARC Australian Laureate Fellowship (FL170100006).
\bibliographystyle{named}
\bibliography{baysOptRevBibV3}

\end{document}